\documentclass[10pt,twocolumn,letterpaper]{article}

\usepackage{cvpr}
\usepackage{times}
\usepackage{amsmath}
\usepackage{amssymb}
\usepackage[misc]{ifsym}
\usepackage{graphicx}
\usepackage{color}
\usepackage{enumitem}
\usepackage{nccbbb}
\usepackage{mathrsfs}

\usepackage{algorithm}
\usepackage{algorithmicx}
\usepackage{algpseudocode}
\usepackage{multirow}
\usepackage{booktabs}
\usepackage{array}
\usepackage{bm}

\usepackage[breaklinks=true,bookmarks=false]{hyperref}

\cvprfinalcopy 

\def\cvprPaperID{****} 

\begin{document}

\title{Deep Distance Transform for Tubular Structure Segmentation in CT Scans}

\author{
Yan Wang\textsuperscript{1}\footnotemark[2] \quad 
Xu Wei\textsuperscript{2}\footnotemark[1] \footnotemark[2] \quad
Fengze Liu\textsuperscript{1} \quad
Jieneng Chen\textsuperscript{3}\footnotemark[1] \quad
Yuyin Zhou\textsuperscript{1} \quad
Wei Shen\textsuperscript{1} \quad\\
Elliot K. Fishman\textsuperscript{4} \quad
Alan L. Yuille\textsuperscript{1} \vspace{.3em}\\
\textsuperscript{1}Johns Hopkins University \quad \textsuperscript{2}University of California San Diego
\quad \textsuperscript{3}Tongji University\\
 \textsuperscript{4}The Johns Hopkins University School of Medicine
\vspace{-.5em}
}

\maketitle
 \renewcommand*{\thefootnote}{\fnsymbol{footnote}}
 \setcounter{footnote}{1}
 \footnotetext{This work was done when X. Wei and J. Chen did internship at JHU.}
 \renewcommand*{\thefootnote}{\arabic{footnote}}
 \renewcommand*{\thefootnote}{\fnsymbol{footnote}}
 \setcounter{footnote}{2}
 \footnotetext{Equal Contribution.}
 \renewcommand*{\thefootnote}{\arabic{footnote}}

\begin{abstract}
Tubular structure segmentation in medical images, e.g., segmenting vessels in CT scans, serves as a vital step in the use of computers to aid in screening early stages of related diseases. But automatic tubular structure segmentation in CT scans is a challenging problem, due to issues such as poor contrast, noise and complicated background. A tubular structure usually has a cylinder-like shape which can be well represented by its skeleton and cross-sectional radii (scales). Inspired by this, we propose a geometry-aware tubular structure segmentation method, Deep Distance Transform (DDT), which combines intuitions from the classical {\textbf{distance transform}} for skeletonization and modern deep segmentation networks. DDT first learns a multi-task network to predict a segmentation mask for a tubular structure and a distance map. Each value in the map represents the distance from each tubular structure voxel to the tubular structure surface. Then the segmentation mask is refined by leveraging the shape prior reconstructed from the distance map. We apply our DDT on six medical image datasets. The experiments show that (1) DDT can boost tubular structure segmentation performance significantly ({{e.g.}}, over 13\% improvement measured by DSC for pancreatic duct segmentation), and (2) DDT additionally provides a geometrical measurement for a tubular structure, which is important for clinical diagnosis (e.g., the cross-sectional scale of a pancreatic duct can be an indicator for pancreatic cancer).
\end{abstract}

\section{Introduction}
\label{sec:Introduction}
Tubular structures are ubiquitous throughout the human body, with notable examples including blood vessels, pancreatic duct and urinary tract. They occur in specific environments at the boundary of liquids, solids or air and surrounding tissues, and play a prominent role in sustaining physiological functions of the human body.

In this paper, we investigate automatic tubular organ/tissue segmentation from CT scans, which is important for the characterization of various diseases~\cite{Grelard2017}. For example, pancreatic duct dilatation or abrupt pancreatic duct caliber change signifies high risk for pancreatic ductal adenocarcinoma (PDAC), which is the third most common cause of cancer death in the US~\cite{Chu2017}. Another example is that obstructed vessels lead to coronary heart disease, which is the leading cause of death in the US~\cite{Rosamond2008}.

Segmenting tubular organs/tissues from CT scans is a popular but challenging problem. Existing methods addressing this problem can be roughly categorized into two groups: (1) Geometry-based methods, which build deformable shape models to fit tubular structures by exploiting their geometrical properties~\cite{Wink2000,Yim2001,Antiga2003,Nain2004}, {\em{e.g.}}, a tubular structure can be well represented by its {\em{skeleton}}, aka {\em{symmetry axis}} or {\em{medial axis}}, and it has a cylindrical surface. But, due to the lack of powerful learning models, these methods cannot deal with poor contrast, noise and complicated background. (2) Learning-based methods, which learn a per-pixel classification model to detect tubular structures. The performance of this type of methods is largely boosted by deep learning, especially fully convolutional networks (FCN)~\cite{Long2015,Zhou2017}. FCN and its variants have become out-of-the-box models for tubular organ/tissue segmentation and achieve state-of-the-art results~\cite{Merkow2016,Zhou2019}. However, these networks simply try to learn a class label per voxel, which inevitably ignores the geometric arrangement of the voxels in a tubular structure, and consequently can not guarantee that the obtained segmentation has the right shape.

\begin{figure}[t]
\begin{center}
    \includegraphics[width=0.6\linewidth]{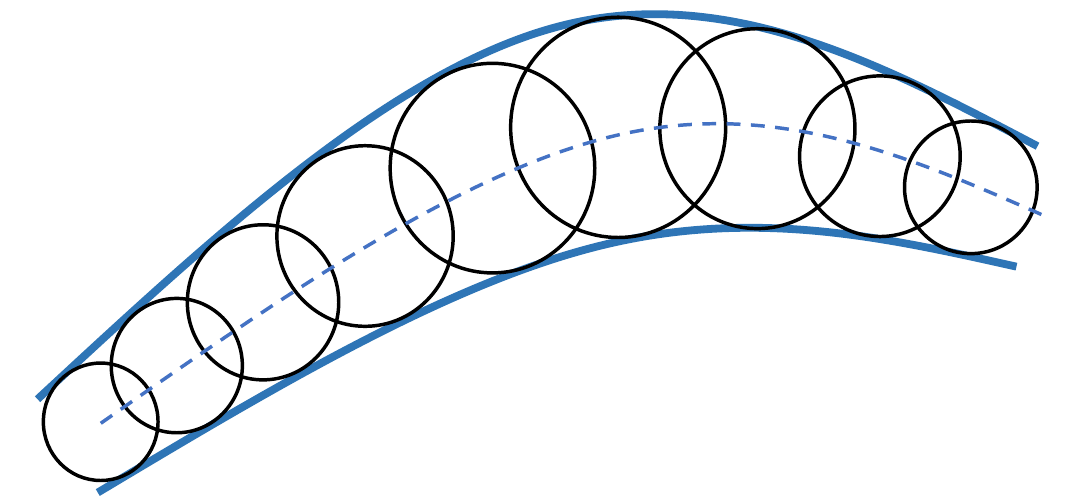}
\end{center}
\vspace{-1ex}
\caption{A tubular shape is presented as the envelope of a family of spheres with continuously changing center points and radii \cite{Benmansour2011}.} 
\vspace{-0.4cm}
\label{Fig:tubular_shape}
\end{figure}

Since a tubular structure can be well represented by its skeleton and the cross-sectional radius of each skeleton point, as shown in Fig.~\ref{Fig:tubular_shape}, these intrinsic geometric characteristics should be taken into account to serve as a valuable prior. To this end, a straightforward strategy is to first train a learning model, {\em{e.g.}}, a deep network, to directly predict whether each voxel is on the skeleton of the tubular structure or not as well as the cross-sectional radius of each skeleton point, and then reconstruct the segmentation of the tubular structure from its skeleton and radii~\cite{Sironi2014}. However, such a strategy has severe limitations: (1) The ground-truth skeletons used for training are not easily obtained. Although they can be approximately computed from the ground-truth segmentation mask by 3D skeletonization methods, skeleton extraction from 3D mesh representation itself is a hard and unsolved problem \cite{Au2008}. Without reliable skeleton ground-truths, the performance of tubular structure segmentation cannot be guaranteed. (2) It is hard for the classifier to distinguish voxels on the skeleton itself from those immediately next to it, as they have similar features but different labels.

To tackle the obstacles mentioned above, we propose to perform tubular structure segmentation by training a multi-task deep network to predict not only a segmentation mask for a tubular structure, but also a {\em{distance map}}, consisting of the distance transform value from each tubular structure voxel to the tubular structure surface, rather than a single skeleton/non-skeleton label. {\em{Distance transform}}~\cite{Rosenfeld1968} is a classical image processing operator to produce a distance map with the same size of the input image, each value in which is the distance from each foreground pixel/voxel to the foreground boundary. Distance transform is also known as the basis of one type of skeletonization algorithms~\cite{Ge1996}, {\em{i.e.}}, the ridge of the distance map is the skeleton. Thus, the predicted distance map encodes the geometric characteristics of the tubular structure. This motivated us to design a geometry-aware approach to refine the output segmentation mask by leveraging the shape prior reconstructed from the distance map. Essentially, our approach performs tubular structure segmentation by an implicit skeletonization-reconstruction procedure with no requirements for skeleton ground-truths. We stress that the distance transform brings two benefits for our approach: (1) Distance transform values are defined on each voxel inside a tubular structure, which eliminates the problem of the discontinuity between the skeleton and its surrounding voxels; (2) distance transform values on the skeleton (the ridge of the distance map) are exactly the cross-sectional radii (scales) of the tubular structure, which is an important {\textbf{geometrical measurement}}. To make the distance transform value prediction more precise, we additionally propose a distance loss term used for network training, which indicates a penalty when predicted distance transform value is far away from its ground-truth. 

We term our method {\em{Deep Distance Transform}} (DDT), as it naturally combines intuitions from the classical distance transform for skeletonization and modern deep segmentation networks. We emphasize that DDT has two advantages over vanilla segmentation networks: (1) It guides tubular structure segmentation by taking the geometric property of tubular structures into account. This reduces the difficulty to segment tubular structures from complex surrounding structures and ensures that the segmentation results have a proper shape prototype; (2) It predicts the cross-sectional scales of a tubular structure as by-products, which are important for the further study of the tubular structure, such as clinical diagnosis and virtual endoscopy~\cite{Bauer2008}.

We verify DDT on six datasets, including five datasets for segmentation task, and one dataset for clinical diagnosis. For segmentation task, the performance of our DDT exceeds all backbone networks by a large margin,  with even over 13\% improvement in terms of Dice-S{\o}rensen coefficient for pancreatic duct segmentation on the famous 3D-Unet \cite{Cicek2016}. The ablation study further shows the effectiveness of each proposed module in DDT. The experiment for clinical diagnosis leverages dilated pancreatic duct as cue for finding missing PDAC tumors by original deep networks, which verifies the potential of our DDT for early diagnosis of pancreatic cancer.

\begin{figure*}[t]
\begin{center}
    \includegraphics[width=1\linewidth]{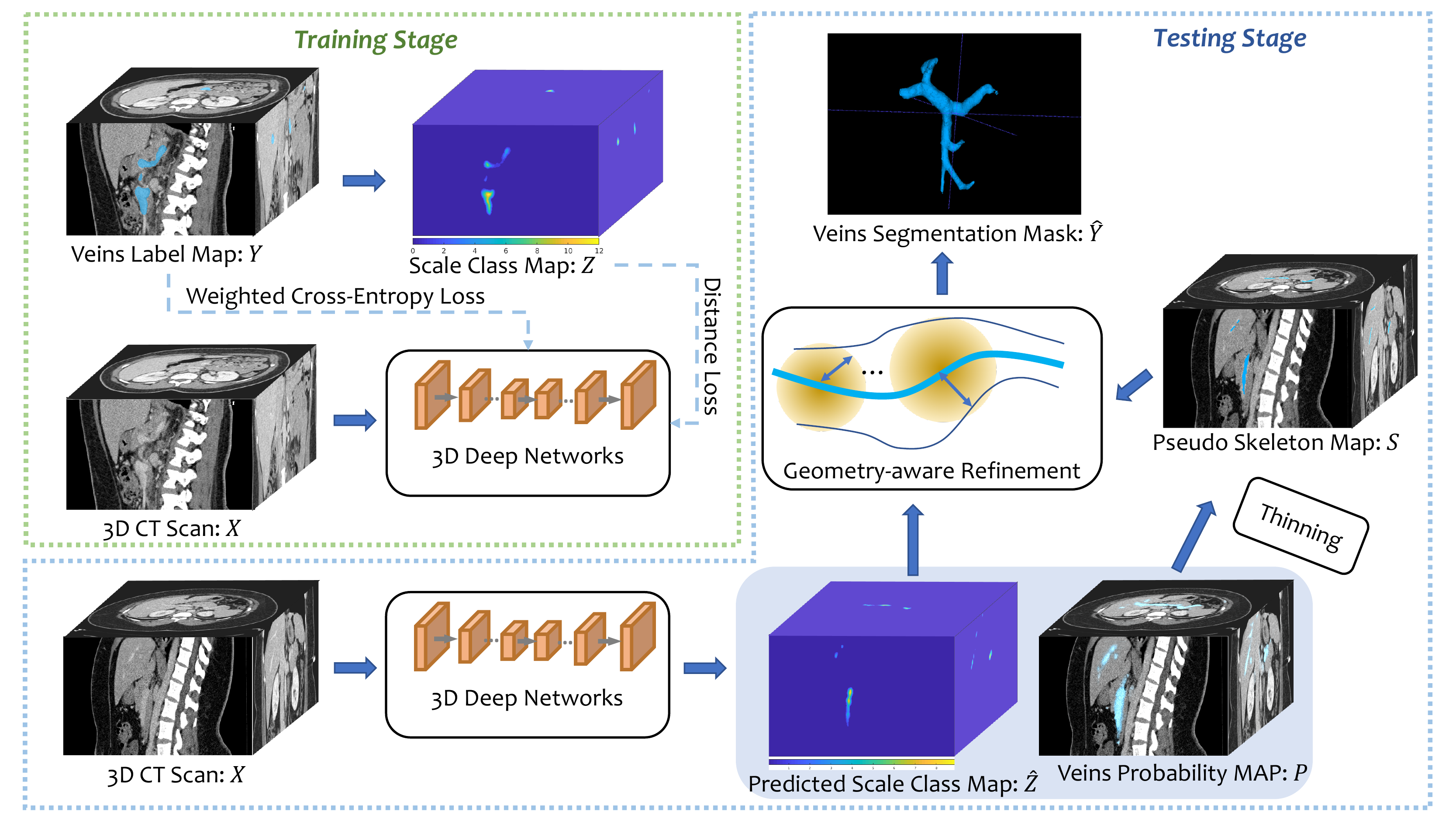}
\end{center}
\vspace{-1ex}
\caption{The training and testing stage of DDT, illustrated on an example of veins segmentation. Our DDT has two head branches: the first one is targeting on the ground-truth label map, which performs per-voxel veins/non-veins classification, and the second head branch is targeting on the scale class map, which performs scale prediction for veins voxels. Then a geometry-aware refinement approach is proposed to leverage the shape prior obtained from the scale class map and the pseudo skeleton map to refine the segmentation mask.}
\label{Fig:Framework}
\end{figure*}

\section{Related Work}
\subsection{Tubular Structure Segmentation}
\subsubsection{Geometry-based Methods}
Various methods have been proposed to improve the performance of tubular structure segmentation by considering the geometric characteristics, and a non-exhaustive overview is given here. (1) Contour-based methods extracted the segmentation mask of a tubular structure by means of approximating its shape in the cross-sectional domain \cite{Alvarez2017,Caselles1997}. 
(2) Minimal path approaches conducted tubular structure tracking and were usually interactive. They captured the global minimum curve (energy weighted by the image potential) between two points given by the user \cite{Benmansour2011}. 
(3) Model-based tracking methods required to refine a tubular structure model, which most of the time adopted a 3D cylinder with elliptical or circular section. At each tracking step, they calculated the new model position by finding the best model match among all possible new model positions \cite{Bauer2010}. 
(4) Centerline based methods found the centerline and estimated the radius of linear structures. For example, multiscale centerline detection method proposed in \cite{Sironi2014} adopted the idea of distance transform, and reformulated centerline detection and radius estimation in terms of a regression problem in 2D. Our work fully leverages the geometric information of a tubular structure, proposing a distance transform algorithm to implicitly learn the skeleton and cross-sectional radius, and the final segmentation mask is reconstructed by adopting the shape prior of the tubular structure\vspace{-0.6em}.

\subsubsection{Learning-based Method}
Learning-based method for tubular structure segmentation infers a rule from labeled training pairs, one for each pixel. Traditional methods such as 2-D Gabor wavelet and classifier combination \cite{Soares2006}, ridge-based segmentation \cite{Staal2004}, and random decision forest based method \cite{Annunziata2015} achieved considerable progress.
In the past years, various 2D and 3D deep segmentation networks have become very popular. Some multi-organ segmentation methods \cite{Wang2019,Roth2017} were proposed to segment multiple organs simultaneously, including tubular organs. DeepVessel \cite{Fu2016} put a four-stage HED-like CNN and conditional random field into an integrated deep network to segment retinal vessel. Kid-Net \cite{Taha2018}, inspired from 3D-Unet \cite{Cicek2016}, was a two-phase 3D network for kidney vessels segmentation.
ResDSN \cite{Zhu2018,Zhu2019} and 3D-Unet \cite{Cicek2016} were used in Hyper-pairing network \cite{Zhou2019} to segment tissues in pancreas including duct by combining information from dual-phase imaging. Besides, 3D-HED and its variant were applied for vascular boundary detection \cite{Merkow2016}. Other scenarios such as using synthetic data to improve endotracheal tube segmentation \cite{Frid-Adar2019}. Cross-modality domain adaptation framework with adversarial learning which dealt with the domain shift in segmenting biomedical images including ascending aorta was also proposed \cite{Dou2018}.

More powerful deep network architecture can produce better segmentation results. But how to leverage geometric information while employing the power of deep network is a more intriguing problem, especially for tubular structure. Our work aims at designing an integrated framework which mines traditional distance transform and model deep networks for such cylinder-like shape structure, which is not studied in prior research.

\subsection{Learning-based Skeleton Extraction}
Learning-based skeleton extraction from natural images has been widely studied in recent decades \cite{Tsogkas2012,Shen2016,Sironi2014,Levinshtein2013,Lee2013} and achieved promising progress with the help of deep learning \cite{Shen2017,Ke2017,Zhao2018,Wang2019_deepflux}. Shen {\em{et al.}} \cite{Shen2017} showed that multi-task learning, {\em{i.e.}}, jointly learning skeleton pixel classification and skeleton scale regression, was important to obtain accurate predicted scales, and it was useful for skeleton-based object segmentation. 

However, these methods cannot be directly applied to tubular structure segmentation, since they require the skeleton ground-truth, which is not easy to obtain from a 3D mask due to the commonly existed annotation errors for medical images \cite{Wang2018}. 

\section{Methodology}
\label{DDT}
We first define a 3D volume $X$ of size $L\times W\times H$ as a function on the coordinate set $V=\{\mathbf{v}|\mathbf{v}\in N_L\times N_W \times N_H \}$, {\em{i.e.}}, $X:V\rightarrow R\subset\mathbb{R}$ where the value on position $\mathbf{v}$ is defined as $x_\mathbf{v}=X(\mathbf{v})$. $N_L$, $N_W$, $N_H$ represent for the integer set ranging from $1$ to $L$, $W$, $H$ respectively, so that the Cartesian product of them can form the coordinate set. Given a 3D CT scan $X$, the goal of tubular structure segmentation is to predict the label $\hat{Y}$ of all voxels in the CT scan, where $\hat{y}_\mathbf{v}\in\{0,1\}$ denotes the predicted label for each voxel at position $\mathbf{v}$, {\em{i.e.}}, if the voxel at $\mathbf{v}$ is predicted as a tubular structure voxel, then $\hat{y}_\mathbf{v}=1$, otherwise $\hat{y}_\mathbf{v}=0$. We also use $\mathbf{v}$ to denote the voxel at position $\mathbf{v}$ in the remaining of the paper for convenience sake. Fig.~\ref{Fig:Framework} illustrates our tubular structure segmentation network, {\em{i.e.}}, DDT. 

\subsection{Distance Transform for Tubular Structure}
\label{DDT:DistanceTransform}
In this section, we discuss how to perform distance transform for tubular structure voxels. Given the ground-truth label map ${Y}$ of the CT scan $X$ in the training phase, let $C_V$ be the set of voxels on the tubular structure surface, which can be defined by
\begin{align}
C_V = \{\mathbf{v}|\ y_{\mathbf{v}}=1, \exists\ \mathbf{u} \in \mathcal{N}(\mathbf{v}), y_\mathbf{u}=0\},
\end{align}
where $\mathcal{N}(\mathbf{v})$ denotes the 6-neighbour voxels of $\mathbf{v}$. Then, by performing distance transform on the CT scan $X$, the distance map $D$ is computed by 
\begin{equation}
d_\mathbf{v}=\left\{
\begin{aligned}
&\min_{\mathbf{u} \in C_V} \|\mathbf{v}-\mathbf{u}\|_2,   &\text{if} \; y_\mathbf{v}=1\\
&0, &\text{if} \; y_\mathbf{v}=0 \\
\end{aligned}
\right..
\end{equation}
Note that, for each tubular structure voxel $\mathbf{v}$, the distance transform assigns it a distance transform value which is the nearest distance from $\mathbf{v}$ to the tubular structure surface $C_V$. Here we use Euclidean distance, as skeletons from Euclidean distance maps are robust to rotations~\cite{Arcelli1992}. 

We further quantize each $d_\mathbf{v}$ into one of $K$ bins by rounding $d_\mathbf{v}$ to the nearest integer, which converts the continuous distance map $D$ to a discrete quantized distance map $Z$, where $z_\mathbf{v} \in \{0,\ldots,K\}$. 
We do this quantization, because training a deep network directly for regression is relatively unstable, since outliers, {\em{i.e.}}, the commonly existed annotation errors for medical images~\cite{Wang2018}, cause a large error term, which makes it difficult for the network to converge and leads to unstable predictions~\cite{RotheTG18}. Based on quantization, we rephrase the distance prediction problem as a classification problem, {\em{i.e.}}, to determine the corresponding bin for each quantized distance. We term the $K$ bins of the quantized distances as $K$ scale classes. We use the term scale since the distance transform values at the skeleton voxels of a tubular structure are its cross-sectional scales. 

\subsection{Network Training for Deep Distance Transform}
\label{DDT:DistanceLoss}
Given a 3D CT scan $X$ and its ground-truth label map $Y$, we can compute its {\em{scale class map}} (quantized distance map) $Z$ according to the method given in Sec.~\ref{DDT:DistanceTransform}. In this section, we describe how to train a deep network for tubular structure segmentation by targeting on both $Y$ and $Z$.

As shown in Fig.~\ref{Fig:Framework}, our DDT model has two head branches. The first one is targeting on the ground-truth label map $Y$, which performs per-voxel classification for semantic segmentation with a weighted cross-entropy loss function $\mathcal{L}_{\text{cls}}$:
\begin{align}
\mathcal{L}_{\text{cls}} &= -\sum_{\mathbf{v}\in V}\Big(\beta_p y_\mathbf{v}\log p_\mathbf{v}(\mathbf{W}, \mathbf{w}_{\text{cls}})\nonumber\\ 
&+\beta_n(1-y_\mathbf{v})\log\big(1-p_\mathbf{v}(\mathbf{W}, \mathbf{w}_{\text{cls}})\big)\Big),
\vspace{-1em}
\end{align}
where $\mathbf{W}$ is the parameters of the network backbone, $\mathbf{w}_{\text{cls}}$ is the parameters of this head branch and $p_\mathbf{v}(\mathbf{W}, \mathbf{w}_{\text{cls}})$ is the probability that $\mathbf{v}$ is a tubular structure voxel as predicted by this head branch. $\beta_p=\frac{0.5}{\sum_\mathbf{v}y_\mathbf{v}}$ and $\beta_n=\frac{0.5}{\sum_\mathbf{v}(1-y_\mathbf{v})}$ are loss weights for tubular structure and background classes respectively. 

The second head branch is predicting on the scale class map $Z$, which performs scale prediction for tubular structure voxels ({\em{i.e.}}, $z_\mathbf{v}>0$). We introduce a new distance loss function $\mathcal{L}_{\text{dis}}$ to learn this head branch:
\begin{align}\label{eq:dis_loss}
\mathcal{L}_{\text{dis}} &= -\beta_p\sum_{\mathbf{v}\in V}\sum_{k=1}^{K}\Bigg(\mathbf{1}(z_\mathbf{v}=k)\Big(\log g_\mathbf{v}^k(\mathbf{W}, \mathbf{w}_{\text{dis}})\nonumber \\  &+ \lambda \omega_\mathbf{v}\log  \big(1-\max_l g_\mathbf{v}^l(\mathbf{W}, \mathbf{w}_{\text{dis}})\big) \Big)\Bigg),
\end{align}
where $\mathbf{W}$ is the parameters of the network backbone, $\mathbf{w}_{\text{dis}}$ is the parameters of the second head branch, $\mathbf{1}(\cdot)$ is an indication function, $\lambda$ is a trade-off parameter which balances the two loss terms (we simply set $\lambda=1$ in our implementation), $g_\mathbf{v}^k(\mathbf{W}, \mathbf{w}_{\text{dis}})$ is the probability that the scale of $\mathbf{v}$ belongs to $k$-th scale class and $\omega_\mathbf{v}$ is a normalized weight defined by $\omega_\mathbf{v}=\frac{|\arg\max_lg_\mathbf{v}^l(\mathbf{W}, \mathbf{w}_{\text{dis}})-z_\mathbf{v}|}{K}$.
Note that, the first term of Eq.~\ref{eq:dis_loss} is the standard softmax loss which penalizes the classification error for each scale class equally. The second term of Eq.~\ref{eq:dis_loss} is termed as {\em{distance loss term}}, which penalizes the difference between each predicted scale class ({\em{i.e.}}, $\max_lg_\mathbf{v}^l(\mathbf{W}, \mathbf{w}_{\text{dis}})$) and its ground-truth scale class $z_\mathbf{v}$, where the penalty is controlled by $\omega_\mathbf{v}$. Finally, the loss function for our segmentation network is $\mathcal{L}=\mathcal{L}_{\text{cls}}+\mathcal{L}_{\text{dis}}$
and the optimal network parameters are obtained by
$(\mathbf{W}^{\ast},\mathbf{w}_{\text{cls}}^{\ast}, \mathbf{w}_{\text{dis}}^{\ast})=\arg\min_{\mathbf{W},\mathbf{w}_{\text{cls}},\mathbf{w}_{\text{dis}}}\mathcal{L}$.

\subsection{Geometry-aware Refinement}
\label{DDT:Post-processing}
Given a 3D CT scan $X$ in the testing phase, for each voxel $\mathbf{v}$, our tubular structure segmentation network, DDT, outputs two probabilities, $p_\mathbf{v}(\mathbf{W}^{\ast}, \mathbf{w}_{\text{cls}}^{\ast})$, which is the probability that $\mathbf{v}$ is a tubular structure voxel and $g_\mathbf{v}^k(\mathbf{W}^{\ast}, \mathbf{w}_{\text{dis}}^{\ast})$, which is the probability that the scale of $\mathbf{v}$ belongs to $k$-th scale class. For notational simplicity, we use $p_\mathbf{v}$ and $g_\mathbf{v}^k$ to denote $p_\mathbf{v}(\mathbf{W}^{\ast}, \mathbf{w}_{\text{cls}}^{\ast})$ and $g_\mathbf{v}^k(\mathbf{W}^{\ast}, \mathbf{w}_{\text{dis}}^{\ast})$, respectively, in the rest of the paper. $p_\mathbf{v}$ provides per-voxel tubular structure segmentation, and $g_\mathbf{v}^k$ encodes the geometric characteristics of the tubular structure. We introduce a geometry-aware refinement approach to obtain the final segmentation result by refining $p_\mathbf{v}$ according to $g_\mathbf{v}^k$. This approach is shown in Fig.~\ref{Fig:Framework} and is processed as follows:

\begin{enumerate}[label=\alph*.]
    \item \textbf{Pseudo skeleton generation.} The probability map $P$ is thinned by thresholding it to generate a binary pseudo skeleton map $S$ for the tubular structure. If $p_\mathbf{v}>T^p$, $s_\mathbf{v}=1$; otherwise, $s_\mathbf{v} = 0$, and $T^p$ is the threshold.
    \item \textbf{Shape reconstruction.} For each voxel $\mathbf{v}$, its predicted scale $\hat{z}_\mathbf{v}$ is given by $\hat{z}_\mathbf{v}=\arg\max_kg_\mathbf{v}^k$. It is known that a shape can be reconstructed from its skeleton by enveloping the maximal balls centered at each skeleton point, {\em{e.g.}}, we can obtain the (binary) reconstructed shape of the tubular structure $\tilde{Y}$ from $S$ by $\tilde{y}_\mathbf{v}=1$  if $\mathbf{v}\in\bigcup_{\mathbf{u}\in\{\mathbf{u'}|s_\mathbf{u'}>0\}}B(\mathbf{u},\hat{z}_\mathbf{u})$, where $B(\mathbf{u},\hat{z}_\mathbf{u})$ is a ball centered at $\mathbf{u}$ with radius $\hat{z}_\mathbf{u}$; otherwise $\tilde{y}_\mathbf{v}=0$. However, the predicted scale $\hat{z}_\mathbf{u}$ is quantized, which leads to an non-smooth surface. Therefore, we fit a Gaussian kernel to soften each ball and obtain a soft reconstructed shape $\tilde{Y}^s$ instead: 
    \begin{equation}\label{eq:reconstruction}
    \tilde{y}^s_\mathbf{v} = \sum_{\mathbf{u}\in\{\mathbf{u'}|s_\mathbf{u'}>0\}}c_\mathbf{u}\Phi(\mathbf{v}; \mathbf{u},\bm{\Sigma}_\mathbf{u}),
    \end{equation}
    where $\Phi(\cdot)$ denotes the density function of a multivariate normal distribution, $\mathbf{u}$ is the mean and $\bm{\Sigma}_\mathbf{u}$ is the co-variance matrix. According to the 3-sigma rule, we set $\bm{\Sigma}_\mathbf{u}=(\frac{\hat{z}_\mathbf{u}}{3})^2\bm{I}$, where $\bm{I}$ is an identity matrix. We notice that the peak of $\Phi(\cdot; \mathbf{u},\bm{\Sigma}_\mathbf{u})$ becomes smaller if $\hat{z}_\mathbf{u}$ is larger. To normalize the peak of each normal distribution, we introduce a normalization factor $c_\mathbf{u}=\sqrt{(2\pi)^3 \text{det}(\Sigma_\mathbf{u})}$. 
    \item \textbf{Segmentation refinement.} We use the soft reconstructed shape $\tilde{Y}^s$ to refine the segmentation probability $p_\mathbf{u}$, which results in a refined segmentation map $\tilde{Y}^r$:
    \begin{equation}
    \tilde{y}^r_\mathbf{v} = \sum_{\mathbf{u}\in\{\mathbf{u'}|s_\mathbf{u'}>0\}}p_\mathbf{u}c_\mathbf{u}\Phi(\mathbf{v}; \mathbf{u},\bm{\Sigma}_\mathbf{u}).
    \end{equation}
    The final segmentation mask $\hat{Y}$ is obtained by thresholding $\tilde{Y}^r$, {\em{i.e.}}, if $\tilde{y}^r_\mathbf{v} > T^r$, $\hat{y}_\mathbf{v} = 1$, otherwise, $\hat{y}_\mathbf{v} = 0$, where $\tilde{y}^r_\mathbf{v}$ and $\hat{y}_\mathbf{v}$ are the value of voxel at position $\mathbf{v}$ of $\tilde{Y}^r$ and $\hat{Y}$, respectively.
\end{enumerate}

As mentioned in Sec.~\ref{sec:Introduction}, the predicted scale $\hat{z}_\mathbf{v}$ is a geometrical measurement for a tubular structure, which is essential for clinical diagnosis. We will show one clinical application in Sec.~\ref{Sec:findPDAC}. 

\section{Experiments}
\label{Experiments}
In this section, we conduct the following experiments: we first evaluate our approach on five segmentation datasets, including (1) the dataset used in \cite{Zhou2019}, (2) three tubular structure datasets created by radiologists in our team, and (3) hepatic vessels dataset in Medical Segmentation Decathlon (MSD) challenge \cite{Simpson2019}. Then, as we mentioned in Sec.~\ref{sec:Introduction}, our DDT predicts cross-sectional scales as by-products, which are important for applications such as clinical diagnosis. We show that the cross-sectional scale is an important measurement for predicting the dilation degree of a pancreatic duct, which can help find the PDAC tumors missed in \cite{Zhu2019}, without increasing the false positives.

\subsection{Tubular Structure Segmentation}

\subsubsection{Implementation Details and Evaluation Metric}
\label{subsec:implementation}
Our implementation is based on PyTorch. For data pre-processing, followed by \cite{Zhou2019}, we truncate the raw intensity values within the range of $[-100, 240]$ HU and normalize each CT scan into zero mean and unit variance. Data augmentation ({\em{i.e.}},translation, rotation and flipping) is conducted in all the methods, leading to an augmentation factor of 24. During training, we randomly sample patches of a specified size ({\em{i.e.}}, 64) due to memory issue. We use exponential learning rate decay with $\gamma=0.99$. During testing, we employ the sliding window strategy to obtain the final predictions. The groundtruth distance map for each tubular structure is computed by finding the euclidean distance of each foreground voxel to its nearest boundary voxels. The segmentation accuracy is measured by the well-known Dice-S{\o}rensen coefficient (DSC) in the rest of the paper, unless otherwise specified.

\subsubsection{The PDAC Segmentation Dataset \cite{Zhou2019}}
We first study the PDAC segmentation dataset \cite{Zhou2019} which has 239 patients with pathologically proven PDAC. All CT scans are contrast enhanced images and our experiments are conducted on only portal venous phase. We follow the same setting and the same cross-validation as reported in \cite{Zhou2019}. DSCs for three structures were reported in \cite{Zhou2019}: abnormal pancreas, PDAC mass and pancreatic duct. We only show the average and standard deviation over all cases for pancreatic duct, which is a tubular structure.

\paragraph{Results and Discussions.}
To evaluate the performance of the proposed DDT framework, we compare it with a per-voxel classification method~\cite{Zhou2019}, termed as {\bf{SegBaseline}} in Table~\ref{Tab:Zhou_Results}. 
It can be seen that our approach outperforms the baseline reported in~\cite{Zhou2019} by a large margin. It is also worth mentioning that although our DDT is only tested on venous phase, the performance is comparable with the hyper-paring network~\cite{Zhou2019} ({\em{i.e.}}, Multi-phase HPN), which integrates multi-phase information ({\em{i.e.}}, arterial phase and venous phase). For 3D-UNet, our DDT even outperforms the multi-phase method by more than 13\% in terms of DSC.  

\begin{table}[t!]
\renewcommand\arraystretch{0.95}
\small
\centering
\caption{Performance comparison (DSC, $\%$) on pancreatic duct segmentation (mean $\pm$ standard deviation of all cases). {SegBaseline} stands for per-voxel classification. {Multi-phase HPN} is a hyper-paring network combining CT scans from both \textbf{venous} (V) and \textbf{arterial} (A) phases. Noted that only CT scans in \textbf{venous} phase are used for {SegBaseline} and {DDT}. \textbf{Bold} denotes the best results.}
\label{Tab:Zhou_Results}
\resizebox{1\linewidth}{!}{
\begin{tabular}{lccc}
\toprule[0.15em]
\multirow{2}{*}{Methods} & \multirow{2}{*}{Phase}& \multicolumn{2}{c}{Backbone Networks}\\
\cmidrule(lr){3-4}
 & &3D-UNet & ResDSN \\
\midrule
SegBaseline \cite{Zhou2019}& V& 40.25~$\pm$~27.89 & 49.81~$\pm$~26.23\\
Multi-phase HPN \cite{Zhou2019} & A+V &44.93~$\pm$~24.88 & \textbf{56.77}~$\pm$~\textbf{23.33}\\
DDT (Ours) & V& \textbf{58.20}~$\pm$~\textbf{23.39} & {55.97}~$\pm$~24.76\\
\bottomrule[0.15em]
\end{tabular}
}
\end{table}

\begin{figure}[t]
\vspace{-0.5em}
\begin{center}
    \includegraphics[width=1\linewidth]{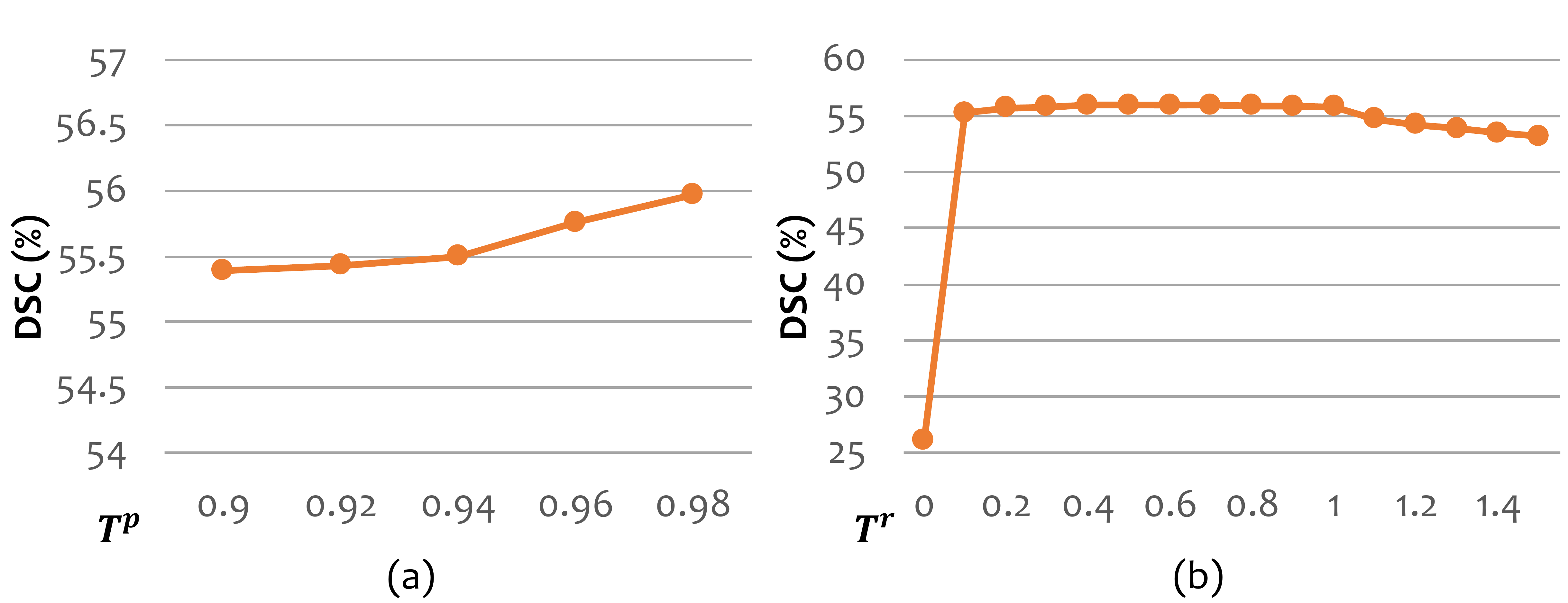}
\end{center}
\vspace{-1ex}
\caption{Performance changes by varying (a) pseudo skeleton generation parameter $T^p$ and (b) segmentation refinement parameter $T^r$.}
\label{Fig:parameter}
\vspace{-0.5em}
\end{figure}

\paragraph{Ablation Study.}
We conduct ablation experiments on the PDAC segmentation dataset, using ResDSN as the backbone. 
These variants of our methods are considered:

\begin{itemize}[leftmargin=*]
\setlength\itemsep{0em}
    \item SegfromSkel: This is the straightforward strategy mentioned in Sec.~\ref{sec:Introduction} for skeleton-based tubular structure segmentation, {\em{i.e.}}, segmenting by reconstructing from the predicted skeleton. The ground-truth skeleton is obtained by the mesh contraction algorithm \cite{Au2008}, and the scale of each skeleton point is defined as its shortest distance to the duct surface. We use the same method in Sec.~\ref{DDT} to instantiate this strategy, but the learning target is the skeleton instead of the duct mask. 
    \item DDT $\lambda=0$, w/o GAR: DDT \textbf{without} distance loss term ($\lambda=0$ in Eq.~\ref{eq:dis_loss}), and \textbf{without} geometry-aware refinement.
    \item DDT $\lambda=0$, w/ GAR: DDT \textbf{without} distance loss term, and \textbf{with} geometry-aware refinement.
    \item DDT $\lambda=1$, w/o GAR: DDT \textbf{with} distance loss term, and \textbf{without} geometry-aware refinement.
    \item DDT $\lambda=1$, w/ GAR: DDT \textbf{with} distance loss term, and \textbf{with} geometry-aware refinement.
\end{itemize}
The results of the ablation experiments are summarized in Table~\ref{Tab:Zhou_Ablation}. 
Then, we aim at discussing parameters in the geometry-aware refinement component. In our implementation, we set $T^p=0.98$ and $T^r=0.5$ in Sec.~\ref{DDT:Post-processing}.
Now we vary each of them and fix the other one to the default value to see how the performance changes. As shown in Fig.~\ref{Fig:parameter}(a), setting a larger $T^p$ leads to better performance. This phenomenon further verifies the advantage of leveraging scale class map to refine the per-voxel segmentation results, {\em{i.e.}}, a thinner pseudo skeleton combined with a scale class map can better represent a tubular structure. Fig.~\ref{Fig:parameter}(b) shows that the performance is not sensitive within the range of $T^r\in[0.1, 1]$\vspace{-0.5em}.

\begin{table}[t!]
\renewcommand\arraystretch{0.95}
\small
\centering
\caption{Ablation study of pancreatic duct segmentation using ResDSN as backbone network. {GAR} indicates the proposed geometry-aware refinement\vspace{0.2em}.}
\label{Tab:Zhou_Ablation}
\begin{tabular}{lc}
\toprule[0.15em]
Method & Average DSC (\%) \\
\midrule[0.1em]
SegBaseline \cite{Zhou2019}& 49.81\\
SegfromSkel & 51.88\\
DDT $\lambda=0$, w/o GAR & 52.73\\
DDT $\lambda=0$, w/  GAR & 54.70\\
DDT $\lambda=1$, w/o GAR & 53.69\\
DDT $\lambda=1$, w/  GAR & \textbf{55.97}\\
\bottomrule[0.15em]
\end{tabular}
\end{table}

\begin{table*}[!tb]
\renewcommand\arraystretch{0.92}
\small
\centering
\caption{Performance comparison (in average DSC, $\%$ and mean surface distance in mm) on three tubular structure datasets by using different backbones. ``$\uparrow$" and ``$\downarrow$" indicate the larger and the smaller the better, respectively. \textbf{Bold} denotes the best results for each tubular structure per measurement.}
\label{Tab:Multi_Tubular}
\begin{tabular}{lccccccc}
\toprule[0.15em]
\multirow{3}{*}{Backbone} & \multirow{3}{*}{Methods} &  \multicolumn{2}{c}{Aorta} & \multicolumn{2}{c}{Veins} & \multicolumn{2}{c}{Pancreatic duct} \\
\cmidrule{3-4} \cmidrule(lr){5-6} \cmidrule(lr){7-8}
& & Average & Mean Surface & Average & Mean Surface & Average &  Mean surface\\ 
& & DSC~$\uparrow$ & Distance~$\downarrow$ & DSC~$\uparrow$ & Distance~$\downarrow$ & DSC~$\uparrow$ & Distance~$\downarrow$ \\
\midrule[0.09em]
\multirow{2}{*}{3D-HED \cite{Merkow2016}}& \footnotesize{SegBaseline} & 90.85 & 1.15 &  73.57 & 5.13 &  46.43 &7.06\\
& DDT & 92.94 & 0.82 & 76.20 & \textbf{3.78} & 54.43 & 4.91\\
\cmidrule(lr){1-8}
\multirow{2}{*}{3D-UNet \cite{Cicek2016}} & \footnotesize{SegBaseline} & 92.01 & 0.94 &  71.57 & 4.46  & 56.63 & 3.64\\
& DDT & \textbf{93.30} & \textbf{0.61} & 75.59 & 4.07 & \textbf{62.31} & \textbf{3.56}\\
\cmidrule(lr){1-8}
\multirow{2}{*}{ResDSN \cite{Zhu2018}} & \footnotesize{SegBaseline} & 89.89 & 1.12 & 71.10 & 6.25 &  55.91 & 4.24 \\
& DDT & 92.57 & 1.10 & \textbf{76.60}& 5.03 & 59.29 & 4.19\\
\bottomrule[0.15em]
\end{tabular}
\vspace{-0.6em}
\end{table*}

\subsubsection{Tubular Structure Datasets}
We then evaluate our algorithm on multiple tubular structures datasets. Radiologists in our team collected 229 abdominal CT scans of normal cases with aorta annotation, 204 normal cases with veins annotation, and 494 abdominal CT scans of biopsy-proven PDAC cases with pancreatic duct annotation. All these three datasets are under IRB approved protocol. 

\begin{figure}[t]
\begin{center}
    \includegraphics[width=1\linewidth]{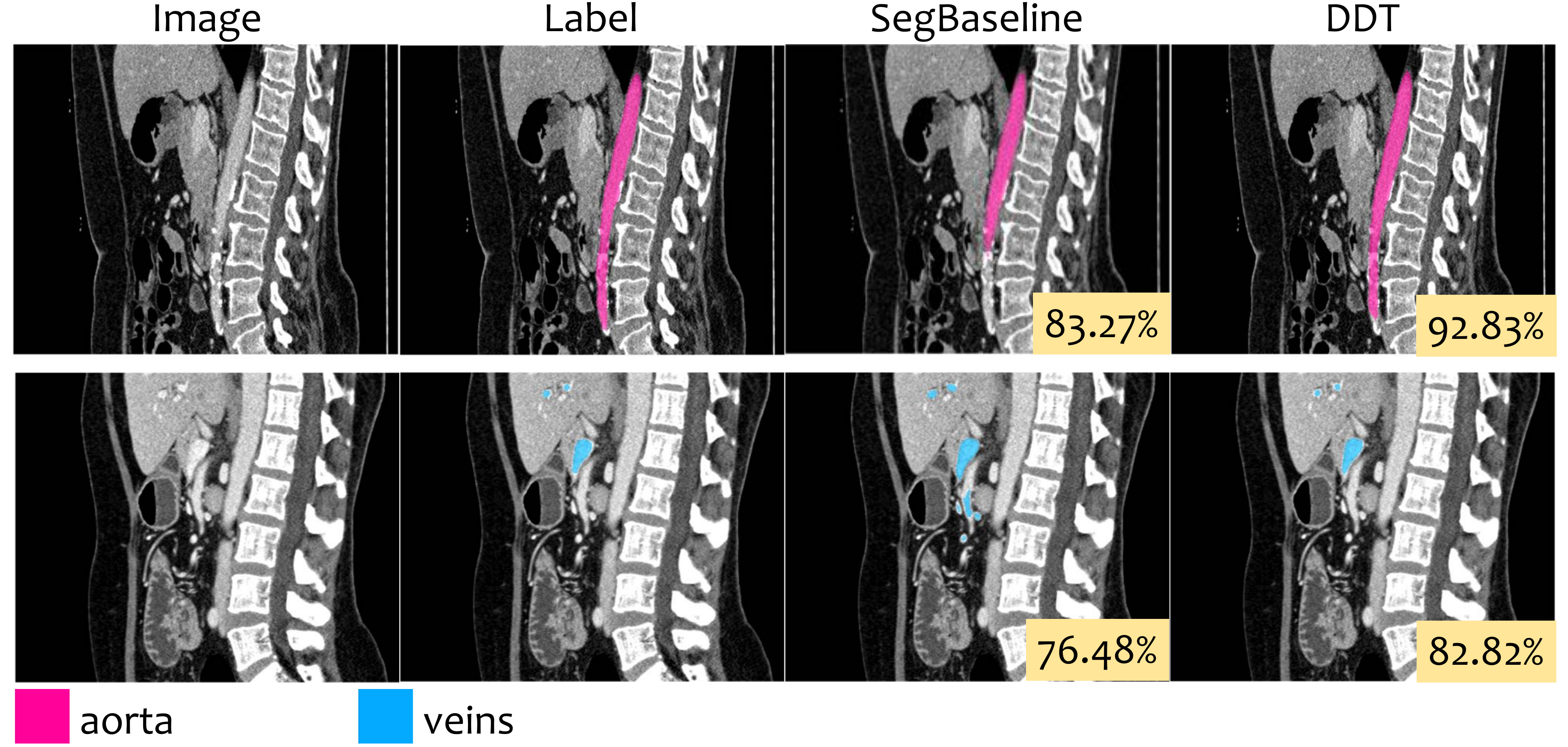}
\end{center}
\vspace{-2ex}
\caption{Illustration of aorta (upper row) and veins (lower row) segmentation results for selected example images. Numbers on the bottom right show segmentation DSCs.
}
\vspace{-0.2em}
\label{Fig:comparison}
\end{figure}

We conduct experiments by comparing our DDT with {SegBaseline} on three backbone networks: 3D-HED \cite{Merkow2016}, 3D-UNet \cite{Cicek2016} and ResDSN \cite{Zhu2018,Zhu2019}. 
The results in terms of DSC and mean surface distance (in mm) are reported in Table~\ref{Tab:Multi_Tubular}. SegBaseline methods on all backbone networks are significantly lower than our approach. In particular, for 3D-HED, DDT outperforms SegBaseline by 8\%, making its strong ability in segmenting small tubular structures like pancreatic duct in medical images. The results are obtained by cross-validation. We also illustrate segmentation results of aorta and veins in Fig.~\ref{Fig:comparison} for qualitative comparison. We can see that compared with SegBaseline, DDT captures geometry information, which is more robust to the noise and complicated background\vspace{-0.5em}.

\subsubsection{Hepatic Vessels Dataset in MSD Challenge\vspace{-0.3em}}
We also test our DDT on a public hepatic vessels dataset in MSD challenge~\cite{Simpson2019}. There are two targets in hepatic vessels dataset: vessels and tumor. As our goal is to segment tubular structure, we aim at vessel segmentation. 
Although this challenge is over, it is still open for submissions. We train our DDT on 303 training cases, and submit vessel predictions of the testing cases to the challenge. 

We simply use ResDSN \cite{Zhu2019} as our backbone network, and follow the same data augmentation as introduced in Sec.~\ref{subsec:implementation}.
We summarize some leading quantitative results reported in the leaderboard in Table~\ref{Tab:MSD}.  This comparison shows the effectiveness of our DDT. 

\subsection{Finding PDAC Tumor by Dilated Duct}
\label{Sec:findPDAC}

\begin{figure}[t]
\begin{center}
    \includegraphics[width=0.9\linewidth]{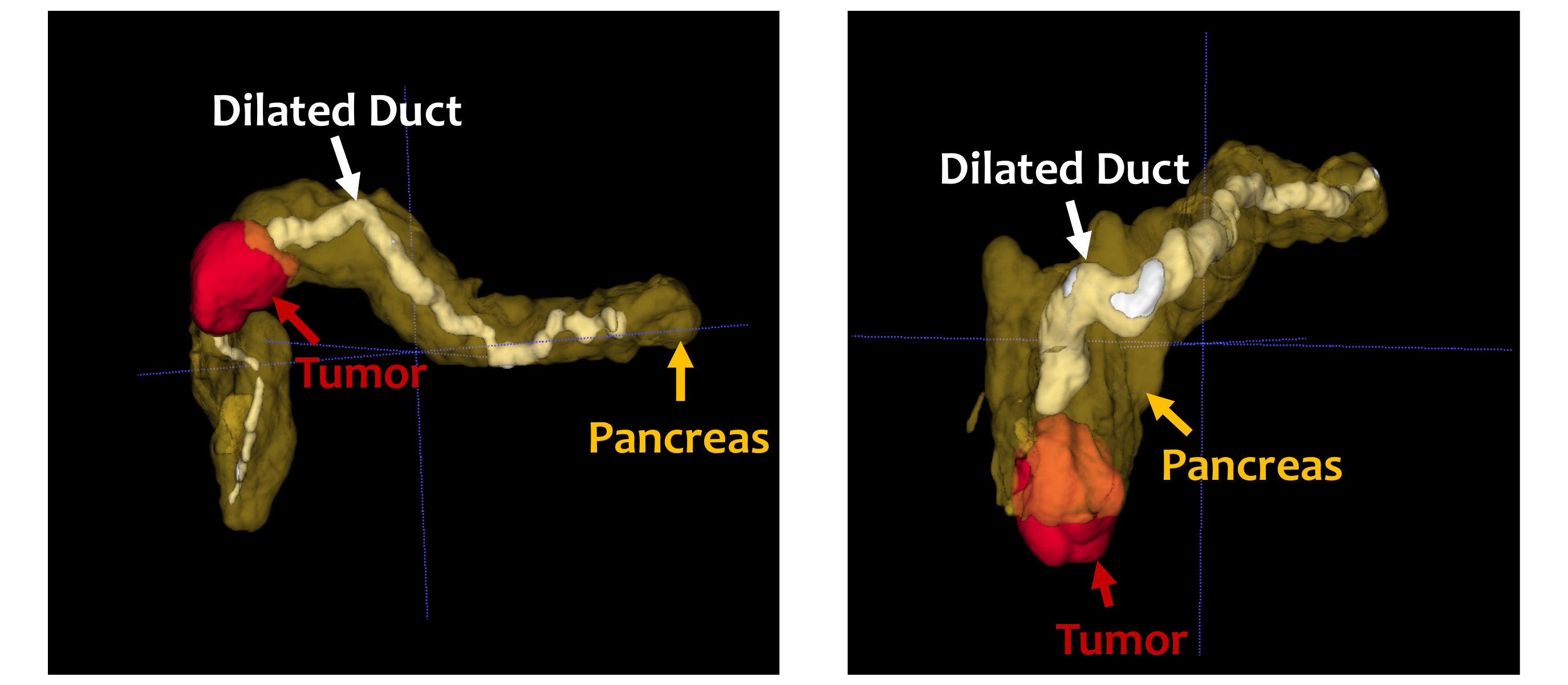}
\end{center}
\vspace{-2ex}
\caption{Examples of PDAC cases. In most PDAC cases, the tumor blocks the duct and causes it to dilate.}

\label{Fig:panc_duct_tumor}
\vspace{-0.9em}
\end{figure}

\begin{table}[t!]
\renewcommand\arraystretch{0.95}
\small
\centering
\caption{Comparison to competing submissions of MSD challenge:~\url{http://medicaldecathlon.com}}
\label{Tab:MSD}
\begin{tabular}{lcc}

\toprule[0.15em]
{Methods} & Average DSC ($\%$) \\
\midrule[0.1em]
DDT (Ours) & \textbf{63.43}  \\
nnU-Net \cite{Isensee2019} & 63.00 \\
UMCT \cite{Xia2019} & 63.00 \\
K.A.V.athlon & 62.00 \\
LS Wang's Group & 55.00 \\
MIMI & 60.00 \\
MPUnet \cite{Perslev2019} & 59.00\\
\bottomrule[0.15em]
\end{tabular}
\vspace{-1em}
\end{table}

\begin{figure*}[t]
\vspace{-1em}
\begin{center}
    \includegraphics[width=0.9\linewidth]{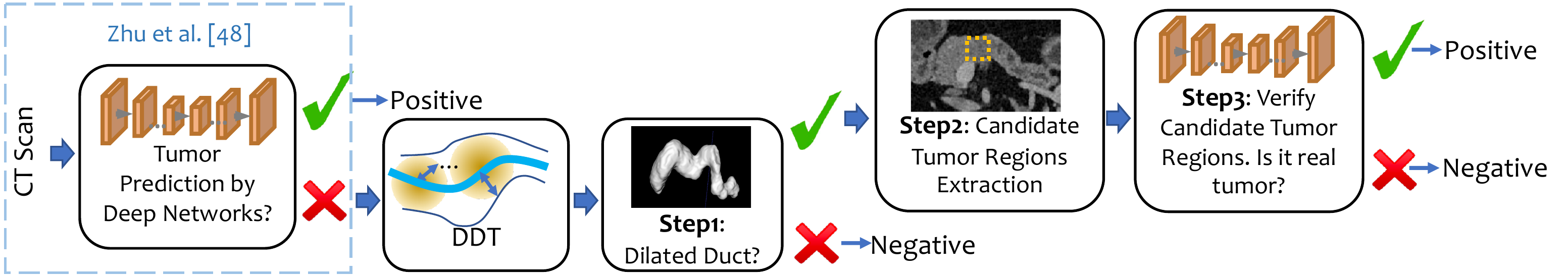}
\end{center}
\vspace{-1ex}
\caption{Flowchart of finding missing PDAC tumor by dilated duct.}
\label{Fig:clinical_flow}
\vspace{-1em}
\end{figure*}
\paragraph{Background.} PDAC is one of the most deadly disease, whose survival is dismal as more than $50\%$ of patients have evidence of metastatic disease at the time of diagnosis. As mentioned in Sec.~\ref{sec:Introduction}, dilated duct is a vital cue for the presence of a PDAC tumor. The reason lies in that in most cases, the tumor blocks the duct and causes it to dilate, as shown in Fig.~\ref{Fig:panc_duct_tumor}. Experienced radiologists usually trace the duct from the pancreas tail onward to see if there exists a truncated duct. If they see the predicted duct pattern as illustrated in Fig.~\ref{Fig:panc_duct_tumor}, they will be alarmed and treat it as a suspicious PDAC case. For computer-aided diagnosis, given a mixture of normal and abnormal CT scans (PDAC cases), if some voxels are segmented as a tumor by a state-of-the-art deep network, we can provide radiologists with tumor locations \cite{Zhu2019}. But, as reported \cite{Zhu2019}, even a state-of-the-art deep network failed to detect 8 PDACs out of 136 abnormal cases. 
As emphasized in \cite{Zhu2019}, for clinical purposes, we shall guarantee a {\em{high sensitivity}} with a reasonable specificity. Then how can we use dilated duct as a cue to help find the PDAC tumor in an abnormal case even if it does NOT have any PDAC tumor prediction by directly applying deep networks\vspace{-0.7em}?

\paragraph{Clinical Workflow.} 

The flowchart of our strategy is illustrated in Fig.~\ref{Fig:clinical_flow}. We apply our DDT on the cases which do not have tumor prediction by \cite{Zhu2019}. Then the predictions of DDT are processed as follows:
\begin{enumerate}[leftmargin=*]
\setlength\itemsep{0em}
    \item \textbf{Find cases with predicted dilated duct}. Let's assume a case has $N$ predicted duct voxels. If $N=0$, then we regard this case as negative. If $N>0$, let's denote the predicted associated scales (radii) are $\{\hat{z}_{\mathbf{v}_i}\}_{i=1}^N$. 
    If $\arg\max_i\hat{z}_{\mathbf{v}_i} > T^s$, {\em{i.e.}}, the largest cross-sectional scale is larger than $T^s$, we regard this is a dilated duct, and a tumor may present on its head location. Otherwise, we treat this case as negative. We set $T^s=3$, since the radius of a dilated duct should be larger than 1.5 mm \cite{Edge07}, and the voxel spatial resolution of the dataset \cite{Zhu2019} is around 0.5 mm$^3$.
    \item \textbf{Extract candidate tumor regions by the location of dilated duct}. We use geodesic distance to find the extreme points of the duct \cite{BaakMBST11}. Then we crop a set of square regions of size $\Re^3$ centered on the extreme points not lying on the tail of the pancreas, since a tumor presenting on the tail of the pancreas will not block a duct. This set of square regions are candidate tumor regions.
    \item \textbf{Verify candidate tumor regions}. As candidate tumor regions may come from both normal and abnormal cases. We should verify whether the candidate region is a real tumor region. From the training set, we randomly crop regions of size $\Re^3$ around PDAC tumor region as positive training data, and randomly crop regions of size $\Re^3$ from normal pancreas as negative training data. Then we train a ResDSN \cite{Zhu2019} to verify these candidate tumor regions. We follow the same criterion used in \cite{Zhu2019} to compute sensitivity and specificity\vspace{-0.8em}.
\end{enumerate}

\paragraph{Experiment Settings.} We follow the same data split as used in \cite{Zhu2019}. We only test our algorithm on the 8 PDAC cases and 197 normal cases which do not have tumor prediction by \cite{Zhu2019}, aiming at finding missing tumor by dilated duct, while not introducing more false positives. $\Re$ is set to be 48 in our experiment\vspace{-0.88em}. 

\begin{figure}[t]
\begin{center}
    \includegraphics[width=1\linewidth]{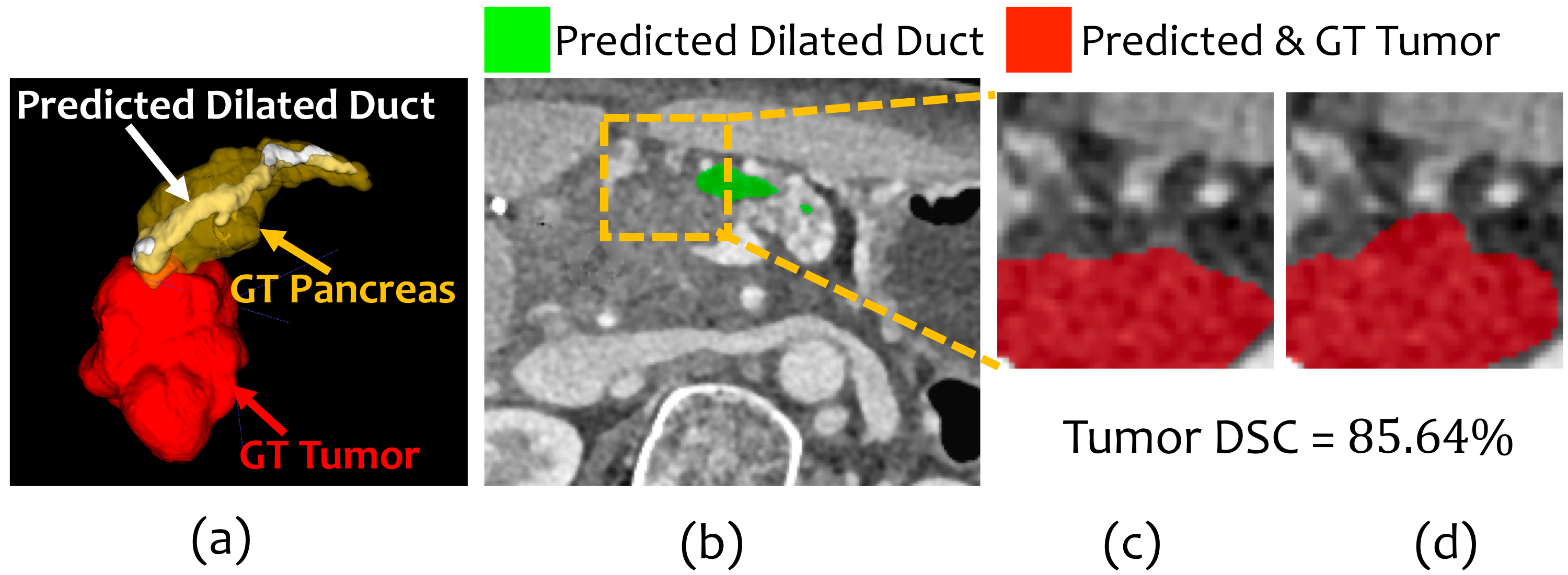}
\end{center}
\vspace{-2ex}
\caption{Examples of finding missed tumor of \cite{Zhu2019} by dilated duct. (a) The ground-truth tumor is right behind one end of the predicted dilated duct. The ground-truth pancreas is shown as a reference. (b) A cropped CT slice with predicted duct (we choose green for better visualization). The yellow dashed box is a candidate tumor region, shown in 2D. (c) and (d) are the same zoomed in image region with predicted and ground-truth tumor, respectively.
}
\label{Fig:find_tumor}
\end{figure}

\paragraph{Analysis.} We compare our results with those of \cite{Zhu2019} in Table~\ref{Tab:Zhu_Results}. In our experiment, 4 out of 8 abnormal cases and 3 out of 197 normal cases have predicted dilated duct by step 1. An example is shown in Fig.~\ref{Fig:find_tumor}(a). The tubular structure residing inside the {\em{ground-truth}} pancreas, right behind the {\em{ground-truth}} tumor is our {\em{predicted}} dilated duct. This leads to overall 18 3D candidate tumor regions by step 2, shown as the yellow dashed box in Fig.~\ref{Fig:find_tumor}(b) visualized in 2D. In step 3, we can successfully find all tumor regions in abnormal cases, and discard non-tumor regions in normal cases. As shown in Fig.~\ref{Fig:find_tumor}(c), our algorithm can find the right tumor, which overlaps with the tumor annotation in Fig.~\ref{Fig:find_tumor}(d). 

It should be emphasized that dilated duct helps us to {\bf{narrow the searching space}} of the tumor, so that we are able to focus on a finer region. Though we train a same network used in \cite{Zhu2019}, half of the missing tumors in \cite{Zhu2019} can be found. In this way, we are imitating how radiologists detect PDAC, {\em{i.e.}}, they can find visible PDAC tumors easily, but for difficult ones, they will seek help from dilated ducts. 

\begin{table}[t!]
\small
\centering
\caption{Normal vs. abnormal classification results. Zhu {\em{et al.}} \cite{Zhu2019} + ours denotes applying our method to find the missing tumor of Zhu {\em{et al.}}.  ``$\uparrow$" and ``$\downarrow$" indicate the larger and the smaller the better, respectively.}
\label{Tab:Zhu_Results}
\resizebox{\columnwidth}{!}{
\begin{tabular}{lccc}
\toprule[0.15em]
{Methods} & {Misses~$\downarrow$} & Sensitivity~$\uparrow$ & Specificity~$\uparrow$\\
\midrule
Zhu {\em{et al.}} \cite{Zhu2019} & 8/136 & 94.1\% & 98.5\%\\
Zhu {\em{et al.}} \cite{Zhu2019} + Ours & 4/136 & 97.1\% & 98.5\%\\
\bottomrule[0.15em]
\end{tabular}
}
\vspace{-0.9em}
\end{table}

\section{Conclusions}
\label{Conclusions}
In this paper, we present Deep Distance Transform (DDT) for accurate tubular structure segmentation, which combines intuitions from the classical distance transform for skeletonization and modern deep segmentation networks. DDT guides tubular structure segmentation by taking the geometric property of the tubular structure into account, which not only leads to a better segmentation result, but also provides the cross-sectional scales, {\em{i.e.}}, a geometric measure for the thickness of tubular structures. We evaluated our approach on six datasets including four tubular structures: pancreatic duct, aorta, veins and vessels. Experiment shows the superiority of the proposed DDT for tubular structure segmentation and clinical application.

{\small
\bibliographystyle{ieee_fullname}
\bibliography{egbib}

\begin{thebibliography}{10}\itemsep=-1pt

\bibitem{Alvarez2017}
Luis {\'{A}}lvarez, Agust{\'{\i}}n Trujillo{-}Pino, Carmelo Cuenca, Esther
  Gonz{\'{a}}lez, Julio Esclar{\'{\i}}n, Luis G{\'{o}}mez, Luis Mazorra, Miguel
  Alem{\'{a}}n{-}Flores, Pablo~G. Tahoces, and Jos{\'{e}}~M.
  Carreira{-}Villamor.
\newblock Tracking the aortic lumen geometry by optimizing the 3d orientation
  of its cross-sections.
\newblock In {\em Proc. MICCAI}, 2017.

\bibitem{Annunziata2015}
Roberto Annunziata, Ahmad Kheirkhah, Pedram Hamrah, and Emanuele Trucco.
\newblock Scale and curvature invariant ridge detector for tortuous and
  fragmented structures.
\newblock In {\em Proc. MICCAI}, pages 588--595, 2015.

\bibitem{Antiga2003}
Luca Antiga, Bogdan Ene{-}Iordache, and Andrea Remuzzi.
\newblock Computational geometry for patient-specific reconstruction and
  meshing of blood vessels from angiography.
\newblock {\em {IEEE} Trans. Med. Imaging}, 22(5):674--684, 2003.

\bibitem{Arcelli1992}
Carlo Arcelli and Gabriella~Sanniti di Baja.
\newblock Ridge points in euclidean distance maps.
\newblock {\em Pattern Recognition Letters}, 13(4):237--243, 1992.

\bibitem{Au2008}
Oscar~Kin{-}Chung Au, Chiew{-}Lan Tai, Hung{-}Kuo Chu, Daniel Cohen{-}Or, and
  Tong{-}Yee Lee.
\newblock Skeleton extraction by mesh contraction.
\newblock {\em {ACM} Trans. Graph.}, 27(3):44:1--44:10, 2008.

\bibitem{BaakMBST11}
Andreas Baak, Meinard M{\"{u}}ller, Gaurav Bharaj, Hans{-}Peter Seidel, and
  Christian Theobalt.
\newblock A data-driven approach for real-time full body pose reconstruction
  from a depth camera.
\newblock In {\em Proc. ICCV}, 2011.

\bibitem{Bauer2008}
Christian Bauer and Horst Bischof.
\newblock Extracting curve skeletons from gray value images for virtual
  endoscopy.
\newblock In {\em International Workshop on Medical Imaging and Virtual
  Reality}, pages 393--402, 2008.

\bibitem{Bauer2010}
Christian Bauer, Thomas Pock, Erich Sorantin, Horst Bischof, and Reinhard
  Beichel.
\newblock Segmentation of interwoven 3d tubular tree structures utilizing shape
  priors and graph cuts.
\newblock {\em Medical Image Analysis}, 14(2):172--184, 2010.

\bibitem{Benmansour2011}
Fethallah Benmansour and Laurent~D. Cohen.
\newblock Tubular structure segmentation based on minimal path method and
  anisotropic enhancement.
\newblock {\em International Journal of Computer Vision}, 92(2):192--210, 2011.

\bibitem{Caselles1997}
Vicent Caselles, Ron Kimmel, and Guillermo Sapiro.
\newblock Geodesic active contours.
\newblock {\em International Journal of Computer Vision}, 22(1):61--79, 1997.

\bibitem{Chu2017}
Linda~C. Chu., Michael~G. Goggins., and Elliot~K. Fishman.
\newblock Diagnosis and detection of pancreatic cancer.
\newblock {\em The Cancer Journal}, 23(6):333--342, 2017.

\bibitem{Cicek2016}
{\"{O}}zg{\"{u}}n {\c{C}}i{\c{c}}ek, Ahmed Abdulkadir, Soeren~S. Lienkamp,
  Thomas Brox, and Olaf Ronneberger.
\newblock 3d u-net: Learning dense volumetric segmentation from sparse
  annotation.
\newblock In {\em Proc. MICCAI}, 2016.

\bibitem{Dou2018}
Qi Dou, Cheng Ouyang, Cheng Chen, Hao Chen, and Pheng{-}Ann Heng.
\newblock Unsupervised cross-modality domain adaptation of convnets for
  biomedical image segmentations with adversarial loss.
\newblock In {\em Proc. IJCAI}, 2018.

\bibitem{Edge07}
Mark~D Edge, Maarouf Hoteit, Amil~P Patel, Xiaoping Wang, Deborah~A Baumgarten,
  and Qiang Cai.
\newblock Clinical significance of main pancreatic duct dilation on computed
  tomography: Single and double duct dilation.
\newblock {\em World J Gastroenterol.}, 13(11):1701--1705, 2007.

\bibitem{Frid-Adar2019}
Maayan Frid{-}Adar, Rula Amer, and Hayit Greenspan.
\newblock Endotracheal tube detection and segmentation in chest radiographs
  using synthetic data.
\newblock In {\em Proc. MICCAI}, pages 784--792, 2019.

\bibitem{Fu2016}
Huazhu Fu, Yanwu Xu, Stephen Lin, Damon Wing~Kee Wong, and Jiang Liu.
\newblock Deepvessel: Retinal vessel segmentation via deep learning and
  conditional random field.
\newblock In {\em Proc. MICCAI}, pages 132--139, 2016.

\bibitem{Ge1996}
Yaorong Ge and J.~Michael Fitzpatrick.
\newblock On the generation of skeletons from discrete euclidean distance maps.
\newblock {\em {IEEE} Trans. Pattern Anal. Mach. Intell.}, 18(11):1055--1066,
  1996.

\bibitem{Grelard2017}
Florent Gr{\'{e}}lard, Fabien Baldacci, Anne Vialard, and Jean{-}Philippe
  Domenger.
\newblock New methods for the geometrical analysis of tubular organs.
\newblock {\em Medical Image Analysis}, 42:89--101, 2017.

\bibitem{Isensee2019}
Fabian Isensee, Jens Petersen, Simon A.~A. Kohl, Paul~F. J{\"{a}}ger, and
  Klaus~H. Maier{-}Hein.
\newblock nnu-net: Breaking the spell on successful medical image segmentation.
\newblock {\em CoRR}, abs/1904.08128, 2019.

\bibitem{Ke2017}
Wei Ke, Jie Chen, Jianbin Jiao, Guoying Zhao, and Qixiang Ye.
\newblock {SRN:} side-output residual network for object symmetry detection in
  the wild.
\newblock In {\em Proc. CVPR}, 2017.

\bibitem{Lee2013}
Tom Sie~Ho Lee, Sanja Fidler, and Sven~J. Dickinson.
\newblock Detecting curved symmetric parts using a deformable disc model.
\newblock In {\em Proc. ICCV}, 2013.

\bibitem{Levinshtein2013}
Alex Levinshtein, Cristian Sminchisescu, and Sven~J. Dickinson.
\newblock Multiscale symmetric part detection and grouping.
\newblock {\em International Journal of Computer Vision}, 104(2):117--134,
  2013.

\bibitem{Long2015}
Jonathan Long, Evan Shelhamer, and Trevor Darrell.
\newblock {Fully Convolutional Networks for Semantic Segmentation}.
\newblock {\em Proc. CVPR}, 2015.

\bibitem{Merkow2016}
Jameson Merkow, Alison Marsden, David~J. Kriegman, and Zhuowen Tu.
\newblock {Dense Volume-to-Volume Vascular Boundary Detection}.
\newblock {\em arXiv preprint arXiv:1605.08401}, 2016.

\bibitem{Nain2004}
Delphine Nain, Anthony~J. Yezzi, and Greg Turk.
\newblock Vessel segmentation using a shape driven flow.
\newblock In {\em Proc. MICCAI}, 2004.

\bibitem{Perslev2019}
Mathias Perslev, Erik~Bj{\o}rnager Dam, Akshay Pai, and Christian Igel.
\newblock One network to segment them all: {A} general, lightweight system for
  accurate 3d medical image segmentation.
\newblock In {\em MICCAI}, 2019.

\bibitem{Rosamond2008}
Wayne Rosamond, Katherine Flegal, Karen Furie, Alan Go, Kurt Greenlund, Nancy
  Haase, Susan~M. Hailpern, Michael Ho, Virginia Howard, Brett Kissela, Steven
  Kittner, Donald Lloyd-Jones, Mary McDermott, James Meigs, Claudia Moy, Graham
  Nichol, Christopher O’Donnell, Veronique Roger, Paul Sorlie, Julia
  Steinberger, Thomas Thom, Matt Wilson, and Yuling Hong.
\newblock Heart disease and stroke statistics-2008 update.
\newblock {\em Circulation}, 117(4), 2008.

\bibitem{Rosenfeld1968}
Azriel Rosenfeld and John~L. Pfaltz.
\newblock Distance functions on digital pictures.
\newblock {\em Pattern Recognition}, 1:33--61, 1968.

\bibitem{Roth2017}
Holger~R. Roth, Hirohisa Oda, Yuichiro Hayashi, Masahiro Oda, Natsuki Shimizu,
  Michitaka Fujiwara, Kazunari Misawa, and Kensaku Mori.
\newblock Hierarchical 3d fully convolutional networks for multi-organ
  segmentation.
\newblock {\em CoRR}, abs/1704.06382, 2017.

\bibitem{RotheTG18}
Rasmus Rothe, Radu Timofte, and Luc~Van Gool.
\newblock Deep expectation of real and apparent age from a single image without
  facial landmarks.
\newblock {\em International Journal of Computer Vision}, 126(2-4):144--157,
  2018.

\bibitem{Shen2016}
Wei Shen, Xiang Bai, Zihao Hu, and Zhijiang Zhang.
\newblock Multiple instance subspace learning via partial random projection
  tree for local reflection symmetry in natural images.
\newblock {\em Pattern Recognition}, 52:306--316, 2016.

\bibitem{Shen2017}
Wei Shen, Kai Zhao, Yuan Jiang, Yan Wang, Xiang Bai, and Alan~L. Yuille.
\newblock Deepskeleton: Learning multi-task scale-associated deep side outputs
  for object skeleton extraction in natural images.
\newblock {\em {IEEE} Trans. Image Processing}, 26(11):5298--5311, 2017.

\bibitem{Simpson2019}
Amber~L. Simpson, Michela Antonelli, Spyridon Bakas, Michel Bilello, Keyvan
  Farahani, Bram van Ginneken, Annette Kopp{-}Schneider, Bennett~A. Landman,
  Geert J.~S. Litjens, Bjoern~H. Menze, Olaf Ronneberger, Ronald~M. Summers,
  Patrick Bilic, Patrick~Ferdinand Christ, Richard K.~G. Do, Marc Gollub,
  Jennifer Golia{-}Pernicka, Stephan Heckers, William~R. Jarnagin, Maureen
  McHugo, Sandy Napel, Eugene Vorontsov, Lena Maier{-}Hein, and M.~Jorge
  Cardoso.
\newblock A large annotated medical image dataset for the development and
  evaluation of segmentation algorithms.
\newblock {\em CoRR}, abs/1902.09063, 2019.

\bibitem{Sironi2014}
Amos Sironi, Vincent Lepetit, and Pascal Fua.
\newblock Multiscale centerline detection by learning a scale-space distance
  transform.
\newblock In {\em CVPR}, 2014.

\bibitem{Soares2006}
Jo{\~{a}}o V.~B. Soares, Jorge J.~G. Leandro, Roberto~M. Cesar, Herbert~F.
  Jelinek, and Michael~J. Cree.
\newblock Retinal vessel segmentation using the 2-d gabor wavelet and
  supervised classification.
\newblock {\em {IEEE} Trans. Med. Imaging}, 25(9):1214--1222, 2006.

\bibitem{Staal2004}
Joes Staal, Michael~D. Abr{\`{a}}moff, Meindert Niemeijer, Max~A. Viergever,
  and Bram van Ginneken.
\newblock Ridge-based vessel segmentation in color images of the retina.
\newblock {\em {IEEE} Trans. Med. Imaging}, 23(4):501--509, 2004.

\bibitem{Taha2018}
Ahmed Taha, Pechin Lo, Junning Li, and Tao Zhao.
\newblock Kid-net: Convolution networks for kidney vessels segmentation from
  ct-volumes.
\newblock In {\em Proc. MICCAI}, 2018.

\bibitem{Tsogkas2012}
Stavros Tsogkas and Iasonas Kokkinos.
\newblock Learning-based symmetry detection in natural images.
\newblock In {\em Proc. ECCV}, pages 41--54, 2012.

\bibitem{Wang2019_deepflux}
Yukang Wang, Yongchao Xu, Stavros Tsogkas, Xiang Bai, Sven~J. Dickinson, and
  Kaleem Siddiqi.
\newblock Deepflux for skeletons in the wild.
\newblock In {\em Proc. CVPR}, 2019.

\bibitem{Wang2019}
Yan Wang, Yuyin Zhou, Wei Shen, Seyoun Park, Elliot~K. Fishman, and Alan~L.
  Yuille.
\newblock Abdominal multi-organ segmentation with organ-attention networks and
  statistical fusion.
\newblock {\em Medical Image Analysis}, 55:88--102, 2019.

\bibitem{Wang2018}
Yan Wang, Yuyin Zhou, Peng Tang, Wei Shen, Elliot~K. Fishman, and Alan~L.
  Yuille.
\newblock Training multi-organ segmentation networks with sample selection by
  relaxed upper confident bound.
\newblock In {\em Proc. MICCAI}, 2018.

\bibitem{Wink2000}
Onno Wink, Wiro~J. Niessen, and Max~A. Viergever.
\newblock Fast delineation and visualization of vessels in 3d angiographic
  images.
\newblock {\em {IEEE} Trans. Med. Imaging}, 19(4):337--346, 2000.

\bibitem{Xia2019}
Yingda Xia, Fengze Liu, Dong Yang, Jinzheng Cai, Lequan Yu, Zhuotun Zhu,
  Daguang Xu, Alan~L. Yuille, and Holger Roth.
\newblock 3d semi-supervised learning with uncertainty-aware multi-view
  co-training.
\newblock In {\em WACV}, 2019.

\bibitem{Yim2001}
Peter~J. Yim, Juan~R. Cebral, Rakesh Mullick, and Peter~L. Choyke.
\newblock Vessel surface reconstruction with a tubular deformable model.
\newblock {\em {IEEE} Trans. Med. Imaging}, 20(12):1411--1421, 2001.

\bibitem{Zhao2018}
Kai Zhao, Wei Shen, Shanghua Gao, Dandan Li, and Ming{-}Ming Cheng.
\newblock Hi-fi: Hierarchical feature integration for skeleton detection.
\newblock In {\em Proc. IJCAI}, 2018.

\bibitem{Zhou2019}
Yuyin Zhou, Yingwei Li, Zhishuai Zhang, Yan Wang, Angtian Wang, Elliot~K
  Fishman, Alan~L Yuille, and Seyoun Park.
\newblock Hyper-pairing network for multi-phase pancreatic ductal
  adenocarcinoma segmentation.
\newblock In {\em Proc.MICCAI}, 2019.

\bibitem{Zhou2017}
Yuyin Zhou, Lingxi Xie, Wei Shen, Yan Wang, Elliot~K. Fishman, and Alan~L.
  Yuille.
\newblock A fixed-point model for pancreas segmentation in abdominal {CT}
  scans.
\newblock In {\em Proc. MICCAI}, 2017.

\bibitem{Zhu2018}
Zhuotun Zhu, Yingda Xia, Wei Shen, Elliot~K. Fishman, and Alan~L. Yuille.
\newblock A 3d coarse-to-fine framework for volumetric medical image
  segmentation.
\newblock In {\em Proc. 3DV}, 2018.

\bibitem{Zhu2019}
Zhuotun Zhu, Yingda Xia, Lingxi Xie, Elliot~K. Fishman, and Alan~L. Yuille.
\newblock Multi-scale coarse-to-fine segmentation for screening pancreatic
  ductal adenocarcinoma.
\newblock In {\em Proc. MICCAI}, 2019.

\end{thebibliography}
}

\end{document}